**Title:** Constrained tensor factorization for computational phenotyping and mortality prediction in patients with cancer

Francisco Y Cai[1], Chengsheng Mao[2], Yuan Luo[2*]

[1]Northwestern University Feinberg School of Medicine, Chicago 60611, IL, USA

[2]Department of Preventive Medicine, Northwestern University Feinberg School of Medicine, Chicago 60611, IL, USA

*corresponding author

**Abstract**

Background: The increasing adoption of electronic health records (EHR) across the US has created troves of computable data, to which machine learning methods have been applied to extract useful insights. In particular, tensor factorization is one such method which has seen much development in recent years. EHR data, represented as a three-dimensional analogue of a matrix (tensor), is decomposed into two-dimensional factors that can be interpreted as computational phenotypes. Constraints imposed during the factorization can promote the discovery of phenotypes with certain desirable properties.

Methods: We apply constrained tensor factorization to derive computational phenotypes and predict mortality in cohorts of patients with breast, prostate, colorectal, or lung cancer in the Northwestern Medicine Enterprise Data Warehouse from 2000 to 2015. In our experiments, we examined using a supervised term in the factorization algorithm, filtering tensor co-occurrences by medical indication, and incorporating additional social determinants of health (SDOH) covariates in the factorization process. We evaluated the resulting computational phenotypes qualitatively and by assessing their ability to predict five-year mortality using the area under the curve (AUC) statistic.

Results: Filtering by medical indication led to more concise and interpretable phenotypes. Mortality prediction performance (AUC) varied under the different experimental conditions and by cancer type (breast: 0.623 – 0.694, prostate: 0.603 – 0.750, colorectal: 0.523 – 0.641, and lung: 0.517 – 0.623). Generally, prediction performance improved with the use of a supervised term and the incorporation of SDOH covariates.

Conclusion: Constrained tensor factorization, applied to sparse EHR data of patients with cancer, can discover computational phenotypes predictive of five-year mortality. The incorporation of SDOH variables into the factorization algorithm is an easy-to-implement and effective way to improve prediction performance.

**Background**

The increasing adoption of electronic health record (EHR) systems across the US has created troves of digital data easily accessible by computing methods [1]. The quantity and variety of data elements is promising for research opportunities but can also be challenging to work with. The data associated with a study cohort may span many dimensions, including demographic information, medications, and diagnoses, and may include structured fields, such as ICD9 billing codes, or unstructured information, such as free text physician notes. To tackle the challenges of extracting insights from EHR data, machine learning algorithms have been applied [2]. Among them, the method of tensor factorization has seen much development in recent years [3-7]. In this area of work, EHR data is represented as a three-dimensional tensor, which is a generalization of a matrix. These dimensions might, for example, represent patients, diagnoses, and medications, with each entry in the tensor representing a three-way interaction, e.g. patient A was prescribed medication B for diagnosis C. Whereas in a two-dimensional matrix, each patient may be associated with a list of diagnoses and medications, a tensor allows for a richer representation: their medications can be associated with some diagnoses and not others. The factorization of the tensor then re-expresses the tensor in terms of lower-dimensional factors that can be interpreted as computational phenotypes.

Recent work has demonstrated the flexibility of this method. In addition to diagnoses and medications, which are frequently found in structured fields, features can also be drawn from unstructured data, such as by applying natural language processing to pathology reports [8]. Gene pathways have also been used as a tensor dimension, with individual genes along another dimension to encode overlapping pathways [5]. At the same time, the objective function that guides the factorization algorithm can be designed to favor the production of computational phenotypes that are distinct and concise, which improves interpretability, or ones that are predictive for specified outcomes, such as mortality or insurance spending [9-11]. In essence, tensor factorization is used as a method of dimensionality reduction and feature discovery. The features outputted – i.e. computational phenotypes – can then be used in a classification task.

In this paper, we apply constrained tensor factorization [10] to EHR data from cancer patients in the Northwestern Medicine Enterprise Data Warehouse (NMEDW) [12] for computational phenotyping. Since constrained tensor factorization was successfully applied by Kim et al. to derive discriminative and distinct phenotypes on the MIMIC-III dataset [13], we reasoned it can also be applied to another set of real-world EHR data. This algorithm extends the standard tensor factorization algorithm, which produces computational phenotypes in an unsupervised manner, by adding a supervised term to their objective function. This term incentivizes the production of phenotypes that are also predictive of patient mortality. Because the supervised term uses logistic regression to guide the derivation of phenotypes, we decided to use a logistic regression model to perform mortality prediction for consistency.

Specifically, our goal was to construct patient-by-medication-by-diagnosis tensors for patient cohorts defined by cancer diagnosis (breast, prostate, colorectal, and lung), and assess the interpretability of the generated computational phenotypes as well as their ability to predict five-year mortality given a one-year observation window. Because the available data did not associate medications with diagnoses with a given patient encounter, we also took the opportunity to compare results based on an equal-correspondence assumption versus using counting only indicated medication-diagnosis pairs. Since demographic information was available, we also modified the tensor factorization algorithm to account for these additional variables during the factorization process. Additionally, while the NMEDW has been used for numerous research studies since its inception in 2007 [12], the use of tensor factorization for cancer computational phenotyping has yet to be evaluated using this data source.

**Methods**

*Data source and processing*

The Northwestern Medicine Enterprise Data Warehouse (NMEDW) is a large central data warehouse, containing more than 2.9 million patients, and serves both clinical and research goals for Northwestern University and a number of clinical partners including Northwestern Memorial Hospital [12]. Patients with a diagnosis of breast, prostate, colorectal, and lung cancer were identified in the NMEDW using ICD9 diagnosis codes (breast: 174.9, 233.0; prostate: 185, 233.4; colorectal: 153, 230.3; lung: 162, 231.2). Four study cohorts were formed, one for each cancer type. Comorbidities and medications recorded at each patient encounter as well as demographic information were also retrieved from the NMEDW. Multiple encounters for the same patient were aggregated. In our dataset, the encounters occurred between 2000 and 2015. An overview of our entire workflow is presented in Figure 1.

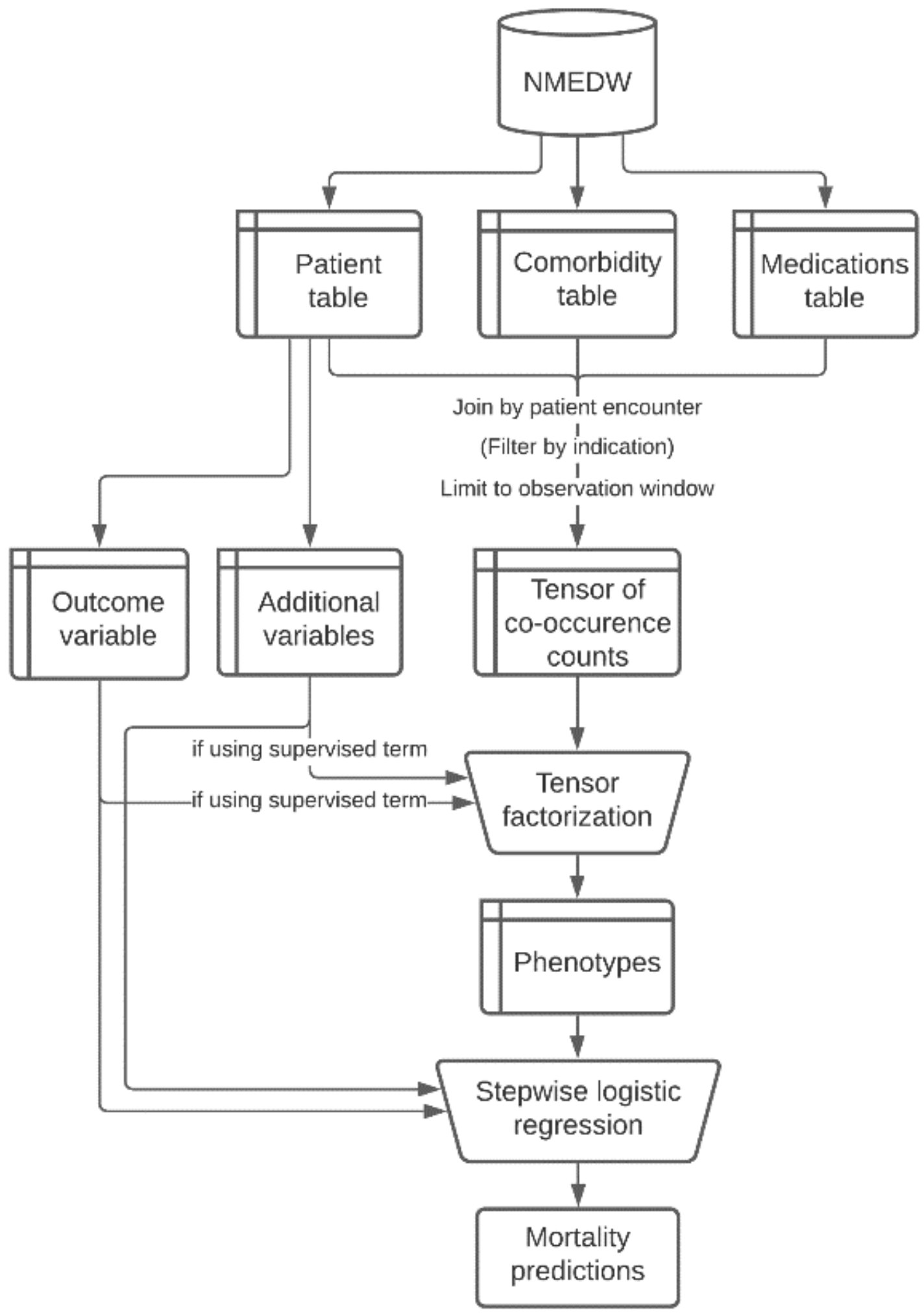

*Figure 1.* **Overview of the workflow.** Patient, comorbidity, and medication information is pulled from the Northwestern Medicine Enterprise Data Warehouse (EDW) and used to create a tensor of co-occurrence counts, which is then factorized to produce phenotype definitions and memberships. Additional variables and an outcome variable participate in factorization if using a supervised term in the objective function, and in the final model selection.

For each patient, we determined the earliest date of cancer diagnosis and if deceased, the date of death. For the latter, we searched both NMEDW and Social Security records. Demographic information included age, sex, race, marital status, insurance status, and zip code. Comorbidity data consisted of ICD9 codes recorded at each patient encounter. Diagnosis codes were retained if they were present in at least 1% of patients. Supplemental V and E codes were excluded, with the exception of a few V codes that may be informative for cancer prognosis, such as estrogen receptor status.

Medication data was compiled from prescriptions, medication administration records, and patient-reported home medications recorded at each patient encounter. Medication names associated with at least 0.5% of patients were retained. In addition, medications approved for all types of cancer [14] were included regardless of prevalence of use in the cohorts. Intravenous fluids, e.g. saline, and medications used for procedures, such as anesthetics, bowel preparations, and contrast agents, were excluded as well. Dose and route of administration were ignored.

Medication names were standardized by mapping synonymous medications to a common generic name. Due to the heterogeneity of naming conventions across different sources of medication information within the NMEDW, this was done manually. The mapping rules, encoded using regular expressions, can be found in the Supplemental Materials. Combination medications (e.g. hydrocodone-acetaminophen) were handled by giving each component its own data entry.

*Tensor Construction*

For each cohort of patients, a tensor with three dimensions – patient, diagnosis, and medication – was constructed for the purpose of computational phenotyping via tensor factorization. Each entry in the tensor is the number of times a specific patient, diagnosis, and medication combination was observed in the same encounter i.e. a count of their co-occurrences. To count co-occurrences, the patient, comorbidity, and medication tables were linked using a unique encounter identifier found in each of the tables. Encounters were limited to those occurring within one year of diagnosis, so we could evaluate mortality prediction using only earlier observations. Co-occurrence counts were truncated at the 99th percentile, similar to Kim et al, to avoid undue influence by frequently administered medications.

While a patient encounter associates a patient with their diagnosis and their medications, there was no correspondence between a medication and the diagnosis for which it was given. Given this, we chose two strategies: using an equal correspondence assumption, where every medication in an encounter is associated with every diagnosis in that encounter, and using indication filtering, where medication-diagnosis co-occurrences are included only if the medication is indicated for that diagnosis. For medication indications, we referred to spreadsheets shared by Vanderbilt University Medical Center [15]. For the few medications not listed, we referred to drug information on the UpToDate database [16]. Indication filtering resulted in fewer patients being represented in the tensor since some patients only had non-indicated co-occurrences. Their medication-diagnosis pairs were reviewed and small number of additional indications were added if a literature search suggested that a medication may have been prescribed for its paired diagnosis. Our additional medication indications can be found in the Supplementary Materials.

*Tensor Factorization*

We evaluated the constrained tensor factorization algorithm developed by Kim et al [10]. Computational phenotyping by tensor factorization takes a three-dimensional tensor as described above, and approximates it using three two-dimensional factor matrices corresponding to the three dimensions of the original tensor, i.e., a patient factor matrix, a diagnosis factor matrix, and a medication factor matrix, with number of rows equal to the number of patients, diagnoses, and medications, respectively. The number of columns is the same across the matrices, custom-defined and denoting the number of phenotypes.

The n-th column from each of the matrices, taken together, can be conceptualized as a "computational phenotype". The n-th column of the diagnosis factor matrix describes the contribution (membership value) of each diagnosis to that phenotype, and similarly for the n-th column of the medication factor matrix. The corresponding column in the patient matrix describes each patient's membership in that phenotype. The membership values are normalized between 0 and 1, and a scalar importance factor is associated with each phenotype, with larger values representing a larger contribution of that phenotype to describing the co-occurrence counts in the tensor.

The tensor factorization method by Kim et al. attempts to produce computational phenotypes that are also able to help discriminate between patients who survived and those who died, if the algorithm is provided that information. Their objective function includes a "supervised" term that quantifies the discriminative ability of the computed phenotypes. Briefly, at every iteration, the current estimates of patient phenotype memberships and the provided mortality labels are used to train a logistic regression model. Those estimated regression coefficients are then used in the next iteration to calculate the supervised term.

We extend their method by allowing additional social determinants of health (SDOH) covariates to be used in the aforementioned logistic regression step. At every iteration, fixed SDOH covariates as well as current estimates of patient phenotype memberships were used to train a logistic regression model for mortality. The estimated regression coefficients for phenotype memberships were then incorporated in the gradient calculation in the next iteration, in the same manner as in Kim et al's algorithm. Previous studies showed that non-clinical patient SDOH covariates also undoubtedly impact individual's outcomes [17, 18]. This is useful when there are additional predictors of mortality that are not diagnoses and medications, and thus cannot be encoded in the tensor. By incorporating these predictors during the supervised tensor factorization, these covariates can be controlled for when deriving patient phenotype memberships. We added 6 additional covariates: indicator variables for male sex, African American race, married marital status, and Medicaid/Medicare insurance, and continuous variables for age at cancer diagnosis and median household income of the patient's zip code. For the categorical variables, missing values were treated as their own level and thus, were included in the "0" value for the indicator variables. Household income data was compiled from 2006-2010 Census Data by the Population Studies Center at University of Michigan and accessed from their website [19].

*Evaluation*

Supervised tensor factorization was performed under different settings characterized by: 1) no indication filtering (i.e. equal correspondence assumption) or indication filtering, 2) the presence or absence of additional SDOH covariates, and 3) the presence or absence of the supervised term in the gradient of the objective function. For each setting, if the supervised term was present, its weight in the gradient equation (referred to as omega by Kim et al.) was positive and chosen to mortality prediction accuracy in the final regression model, as measured by cross-validated area under the curve (see below). Co-occurrences were restricted to those in the observation window, which was within one year of cancer diagnosis, and the outcome was survival at five years post-diagnosis. As with Kim et al's implementation, the number of phenotypes, i.e. the tensor rank, was set at 50. The number of phenotypes selected to the final regression models (see below) was substantially fewer, indicating that having a tensor rank of 50 did not limit our ability to identify informative phenotypes. Unlike Kim et al, we did not include a regularization term for clustering to promote more distinct phenotypes.

The resulting computational phenotypes were reviewed for interpretability. The final estimated patient phenotype memberships, as well as SDOH covariates (if present), were used to fit a logistic regression model using stepwise regression, with an entry and exit significance level of 0.05 and 0.10, respectively. Mortality prediction performance was quantified by the average area under the curve (AUC) across repeated cross-fold validation (10 folds repeated 5 times), with bootstrap 95% confidence intervals.

*Implementation*

Data was extracted from the NMEDW using SQL queries. Data processing and analysis was implemented in R [20] with interactive Jupyter Notebooks [21]. A Matlab implementation of the supervised tensor factorization algorithm was provided by Kim et al. [10] to which our modifications were added. The tensor factorization code with our modifications is available from the corresponding author upon request.

## Results

**Table 1. Tensor characteristics.**

| | Breast | | Prostate | | Colorectal | | Lung | |
|---|---|---|---|---|---|---|---|---|
| | All | Ind. | All | Ind. | All | Ind. | All | Ind. |
| Patients | 3232 | 2319 | 3760 | 2965 | 1217 | 915 | 1701 | 1448 |
| Diagnoses | 149 | 124 | 146 | 109 | 162 | 122 | 215 | 166 |
| Medications | 149 | 129 | 166 | 139 | 184 | 156 | 242 | 213 |
| Dx-med pairs | 7022 | 1208 | 8730 | 1443 | 6807 | 1073 | 11480 | 2152 |
| Mean age at diagnosis | 59.8 | 60.5 | 65.9 | 65.7 | 64.2 | 65.1 | 67.4 | 67.4 |
| Deaths at five years | 332 | 268 | 346 | 269 | 330 | 247 | 794 | 701 |
| Median co-occurrences per pt. | 3 | 2 | 4 | 3 | 6 | 4 | 7 | 5 |
| Total co-occurrences | 58141 | 13999 | 234199 | 46718 | 82918 | 14695 | 149410 | 32852 |

Characteristics of the constructed tensors, organized by cancer type. For each cancer type, two tensors were constructed – one using the equal-correspondence assumption that did not filter out any co-occurrences (columns labelled "all") and one using indication filtering (columns labelled "ind"). The number of distinct diagnosis-medication pairs in each tensor is listed in the "Dx-med pairs" row. Deaths at five years refers to five years post-diagnosis.

*Constructed tensors were sparse*

After data processing, the cohort sizes ranged from 1232 patients with colorectal cancer to 3805 patients with prostate cancer and indication filtering resulted in the removal of 15 - 28% of patients depending on the cohort (Table 1). All diagnoses occurred in the period from 1996 to 2015. The distribution of co-occurrence counts was right-skewed, with the majority of co-occurrences appearing only once. The median number of co-occurrences per patient was few, ranging from 3 to 7 across the different cohorts under the equal-correspondence assumption.

Of the total number of possible diagnosis-medication pairs (number of medications times number of diagnoses), about one-fifth to one-third was actually observed in the same patient encounter. However, many of those pairs were not indicated; for example, since lisinopril was frequently recorded at encounters, it was associated with around 100 diagnoses in the various cohorts. After indication filtering, the vast majority of pairs were removed, resulting in a much smaller total number of co-occurrences and a decrease of 1-2 in the median co-occurrences per patient (Table 1, Figure S1).

*Indication filtering resulted in more concise phenotypes*

**Table 2. Phenotype lengths.**

| | | Breast | | Prostate | | Colorectal | | Lung | |
|---|---|---|---|---|---|---|---|---|---|
| Membership value | Per phenotype | All | Ind. | All | Ind. | All | Ind. | All | Ind. |
| > 0 | Diagnoses | 8.6 | 5.6 | 22.0 | 9.6 | 7.7 | 4.0 | 11.7 | 5.6 |
| | Medications | 32.2 | 11.8 | 45.9 | 16.8 | 42.1 | 11.0 | 55.9 | 19.5 |
| > 0.1 | Diagnoses | 2.3 | 1.4 | 4.5 | 1.8 | 2.5 | 1.3 | 3.1 | 1.4 |
| | Medications | 9.0 | 3.8 | 10.2 | 3.9 | 10.8 | 4.0 | 12.9 | 6.2 |

The average number of diagnoses and medications with non-zero membership values per phenotype, grouped by cancer type. For each cancer type, two columns of results are shown, one from the use of all co-occurrences without indication filtering ("all") and one from the use of indication filtering ("ind"). The rows labeled with membership value > 0.1 shows the average number of diagnoses and medications with a sizable contribution per phenotype.

We define phenotype length as the number of non-zero membership values for a phenotype. When using all co-occurrences, the resulting phenotypes had an average of 8.6 to 21.8 diagnoses, and 32.2 to 55.7 medications with non-zero membership values, depending on the cohort (Table 2). Using indication filtering, the phenotype lengths decreased significantly for all cohorts, to an average of 5.4 to 9.4 diagnoses and 11.1 to 19.5 medications per phenotype. The distribution of membership values for diagnoses was bimodal, with a large peak at values close to 0 and a smaller peak at values close to 1 (Figure S2). The distribution of membership value for medications was unimodal with a peak close to 0. Ignoring components with low membership values (< 0.1), phenotypes had on average 2-4 diagnoses and 9-13 medications without indication filtering, and 1-2 diagnoses and 9-12 medications with indication filtering (Table 2).

Across the different cohorts, the phenotypes with the largest contributions to the tensor decomposition, as defined by the value of lambda in Kim et al. [10], were very similar across cross-validation folds (Figure S3). Those selected by stepwise regression to be in the final model were similar as well (Figure S3).

Two representative sets of phenotypes are shown in Table 3 – one derived with the use of indication filtering and one derived with no indication filtering (i.e. assuming equal correspondence). With no indication filtering, the computed phenotypes contain disparate diagnoses such as diabetes mellitus type 2 and pleural effusion, and some medications in a phenotype do not have a corresponding indicated diagnosis. For example, phenotype 1 with no indication filtering, contains atorvastatin but not hyperlipidemia (Table 3). In contrast, with indication filtering, the diagnoses within each phenotype are more related – phenotype 1 with indication filtering, contains neoplasm of the bone and bone marrow as well as osteoporosis (Table 3). In phenotype 13, chronic disorders including type 2 diabetes mellitus, hypertension, hypercholesterolemia, and coronary atherosclerosis are grouped together. The relatedness of the diagnoses and medications within each phenotype facilitates the annotation of each phenotype with a concise, conceptual label, examples of which can be found in the left-most column of Table 3.

**Table 3. Representative phenotypes from the breast cancer cohort.**

| **No indication filtering** | | |
|---|---|---|
| **Phenotype** | **Diagnoses** | **Medications** |
| 1 | Diabetes mellitus without mention of complication, type II or unspecified type, not stated as uncontrolled, Unspecified pleural effusion, Other respiratory abnormalities | heparin, gabapentin, docusate, acetaminophen, lansoprazole, hydrocodone, furosemide, atorvastatin, venlafaxine, exemestane, insulin, albuterol, hydrocortisone, lorazepam, trazodone, citric acid, sodium citrate, senna, phenylephrine, ciprofloxacin |
| 11 – metatstatic disease | Other malignant neoplasm without specification of site, Secondary malignant neoplasm of bone and bone marrow, Nausea with vomiting, Backache, unspecified | hydromorphone, dexamethasone, docusate, metoclopramide, senna, letrozole, lorazepam, heparin, acetaminophen, hydrocodone, cephalexin, diphenhydramine, ephedrine, lansoprazole, phenylephrine, epinephrine, labetalol, esmolol, tramadol, insulin |
| 12 | Other pulmonary embolism and infarction, Neutropenia, unspecified, Acute kidney failure, unspecified, Calculus of kidney | vancomycin, gabapentin, enoxaparin, oxycodone, famotidine, citric acid, sodium citrate, acetaminophen, ferrous sulfate, prochlorperazine, lansoprazole, ciprofloxacin, metoprolol, insulin, alteplase, diphenhydramine, pegfilgrastim, epinephrine, dexamethasone, esomeprazole |
| 13 | Acute venous embolism and thrombosis of unspecified deep vessels of lower extremity, Other specified cardiac dysrhythmias, Unspecified essential hypertension, Pneumonia, organism unspecified, Other malignant | albuterol, oxycodone, dexamethasone, hydromorphone, acetaminophen, omeprazole, amlodipine, docusate, enoxaparin, metoprolol, megestrol, ketorolac, vancomycin, hydrocodone, senna, zolpidem, heparin, |
| | neoplasm without specification of site | azithromycin, diazepam, clauvanate |
| 16 | Pneumonia, organism unspecified, Unspecified pleural effusion, Fever, unspecified, Acute venous embolism and thrombosis of unspecified deep vessels of lower extremity, Unspecified disorder of kidney and ureter, Edema, Abdominal pain, unspecified site | pantoprazole, hydromorphone, albuterol, vancomycin, furosemide, acetaminophen, hydrocodone, ciprofloxacin, phenylephrine, docusate, senna, tramadol, levofloxacin, famotidine, hydrocortisone, enoxaparin, sertraline, heparin, ketorolac, levothyroxine |
| **With indication filtering** | | |
| 1 - metastatic disease with bone involvement | Secondary malignant neoplasm of bone and bone marrow, Osteoporosis, unspecified, Other malignant neoplasm without specification of site | denosumab, goserelin, hydrochlorothiazide, letrozole, trastuzumab, doxorubicin, darbepoetin, epoetin, tamoxifen, gemcitabine, capecitabine, palbociclib, carboplatin, zoledronate, ciprofloxacin, metoclopramide, propranolol, cyclophosphamide, oxycodone |
| 2 – metastatic disease with neurologic involvement | Secondary malignant neoplasm of brain and spinal cord, Abnormality of gait, Herpes zoster without mention of complication, Enlargement of lymph nodes, Dizziness and giddiness, Headache, Other malignant neoplasm without specification of site | dexamethasone, ciprofloxacin, paclitaxel, letrozole, ibuprofen, hydrocortisone, lapatinib, capecitabine, pamidronate, tamoxifen, erlotinib, doxorubicin, zoledronate, hydrochlorothiazide |
| 9 – generalized metastatic disease | Secondary malignant neoplasm of bone and bone marrow, Secondary malignant neoplasm of brain and spinal cord, Malignant neoplasm of liver, secondary | trastuzumab, pertuzumab, ado-trastuzumab, emtansine, prednisone, gemcitabine, ibuprofen, metoclopramide, dexamethasone, tamoxifen, palonosetron, lapatinib, hydrocodone, fosaprepitant, aprepitant, oxycodone, zoledronate |
| 13 – metabolic disease and other chronic disorders | Diabetes mellitus without mention of complication, type II or unspecified type, not stated as uncontrolled, Unspecified essential hypertension, Pure hypercholesterolemia, Coronary atherosclerosis of unspecified type of vessel, native or graft, Anxiety state, unspecified, Unspecified hereditary and idiopathic peripheral neuropathy | atorvastatin, gabapentin, acetaminophen, insulin, metoclopramide, metformin |
| 27 | Abdominal pain, unspecified site, Edema, Backache, unspecified | hydromorphone, omeprazole, oxycodone, lorazepam, levothyroxine, dexamethasone, prednisone, ciprofloxacin, esomeprazole, lansoprazole, aspirin, prochlorperazine, diphenhydramine |

Shown in this table are the top five phenotypes selected to be in the final model, ranked in order of contribution to the input tensor, for the case where indication filtering was not used, and for the case where it was. Tensor factorization was performed with the supervised term and with SDOH covariates. Where possible, phenotypes were annotated with a more concise label (found in the left-most column).

### *Indication filtering in combination with additional covariates increased prediction accuracy*

After tensor factorization, each patient is described by membership values to each computational phenotype. These memberships are used as features for predicting mortality, and stepwise logistic regression is used to select the final model. The co-occurrences are derived from data within the first year of cancer diagnosis, and mortality is assessed at five years after diagnosis.

We performed experiments comparing the use of indication filtering versus without (i.e. equal correspondence), and the inclusion of additional SDOH covariates versus none. For each set of conditions, we varied the weight of the supervised term in the algorithm's objective function (referred to as omega). A higher value for omega places a greater weight on producing phenotypes that are more discriminative. An omega of 0 means the tensor factorization proceeded without any input from the supervised term. However, the covariates, when used, were still available in the final model selection even if there was no supervised term for them to participate in during the factorization process (Figure 1).

**Table 4. Prediction performance.**

| | All co-occurrences | | Indicated only | |
|---|---|---|---|---|
| Features | Unsupervised | Supervised | Unsupervised | Supervised |
| **Breast cancer** | | | | |
| Covariates only | 0.627<br>(0.610, 0.644) | -- | 0.623<br>(0.609, 0.636) | -- |
| Phenotypes only | 0.639<br>(0.621, 0.658) | 0.652<br>(0.637, 0.666) | 0.654<br>(0.635, 0.673) | 0.649<br>(0.631, 0.664) |
| Phenotypes + covariates | 0.651<br>(0.637, 0.669) | 0.685<br>(0.673, 0.698) | 0.667<br>(0.651, 0.682) | 0.694<br>(0.681, 0.707) |
| **Prostate cancer** | | | | |
| Covariates only | 0.740<br>(0.728, 0.752) | -- | 0.734<br>(0.720, 0.748) | -- |
| Phenotypes only | 0.604<br>(0.590, 0.618) | 0.603<br>(0.586, 0.621) | 0.678<br>(0.659, 0.695) | 0.666<br>(0.649, 0.683) |
| Phenotypes + covariates | 0.740<br>(0.726, 0.753) | 0.733<br>(0.718, 0.747) | 0.740<br>(0.726, 0.757) | 0.750<br>(0.729, 0.768) |
| **Colorectal cancer** | | | | |
| Covariates only | 0.574<br>(0.5458, 0.591) | -- | 0.591<br>(0.575, 0.607) | -- |
| Phenotypes only | 0.578 (0.561, 0.595) | 0.578 (0.562, 0.595) | 0.523 (0.506, 0.542) | 0.565 (0.541, 0.583) |
| Phenotypes + covariates | 0.641 (0.627, 0.656) | 0.638 (0.623, 0.649) | 0.626 (0.612, 0.643) | 0.633 (0.615, 0.649) |
| **Lung cancer** | | | | |
| Covariates only | 0.588 (0.576, 0.600) | -- | 0.579 (0.568, 0.589) | -- |
| Phenotypes only | 0.525 (0.513, 0.538) | 0.558 (0.542, 0.574) | 0.517 (0.502, 0.530) | 0.548 (0.532, 0.563) |
| Phenotypes + covariates | 0.619 (0.611, 0.627) | 0.623 (0.611, 0.639) | 0.601 (0.588, 0.613) | 0.614 (0.600, 0.627) |

The average AUC (and bootstrap 95% confidence intervals) are shown above, grouped by cancer type. The columns are organized by whether all co-occurrences (i.e. no indication filtering) or only indicated co-occurrences were used, and whether a supervised term was included in the objective function. Within each cancer type, the rows are organized by the predictors used. Note: for "covariates only", the SDOH covariates were used in a logistic regression model to establish a baseline; no tensor factorization was performed. Despite the absence of tensor factorization, indication filtering still affected results because the patient cohorts were slightly different due to the exclusion of patients who did not have any indicated co-occurrences. In the case for "phenotypes + covariates", the SDOH covariates were incorporated into the tensor factorization step only under the supervised condition; however, they were always available for the final model selection.

The mean area under the curve (AUC), averaged over 10-fold cross-validation repeated 5 times was used to assess the classification performance of the final logistic regression model. Performance varied between the cohorts; the best results were seen in the prostate cancer cohort and the worst in the lung cancer cohort (Table 4), and generally with the use of phenotype memberships plus covariates, with the supervised term in the objective function.

The addition of covariates to phenotype memberships consistently outperformed phenotype memberships only. In most cases, covariates with phenotype memberships also outperformed models with covariates only, with the exception of the prostate cancer cohort, in which the two performed similarly.

The inclusion of a supervised term improved performance in most cases when indication filtering was used – the exception being breast and prostate cancer when using phenotype memberships only. However, without indication filtering, the supervised term only improved performance in the breast cancer and lung cancer cohorts.

As for indication filtering, its use generally improved performance in the breast and prostate cancer cohorts, for both the unsupervised and supervised cases, with the exception of two experiments where it led to very similar performance (breast cancer with phenotype memberships only and a supervised term, and prostate cancer with covariates and phenotype memberships without a supervised term). For the colorectal and lung cancer cohorts, indication filtering led to worse performance, and the decreased performance was generally more pronounced in the experiments where no supervised term was used.

**Discussion**

We applied a supervised tensor factorization algorithm to a set of EHR data about cancers for computational phenotyping, and used the resulting phenotypes mortality prediction. The algorithm was applied to both inpatient and outpatient encounters of patients with cancer in the Northwestern Medicine Enterprise Data Warehouse [12].

In the course of tensor construction, the lack of an explicit medication-diagnosis correspondence in the EHR data presented an opportunity to compare the two strategies to establish a correspondence post-hoc. We found, as one might expect, that using only indicated correspondences led to more concise phenotypes than using an equal correspondence assumption. Within the phenotypes, the diagnoses were more related to each other, and the medications, by design, had corresponding diagnoses.

Altogether, this resulted in phenotypes that could be more easily conceptualized with a summary label, such as "metastatic disease with bone involvement" or "metabolic disease and other chronic disorders." However, the phenotypes derived after indication filtering were not as discriminative, suggesting that some information may have been lost, although the addition of supervision during factorization rescued the performance of the final regression model to some extent. Our indication filtering based on curated medication indications may have been too restrictive. In recent work by Yin et al., the correspondence was inferred simultaneously with tensor factorization, leading to improved classification performance [22].

The algorithm's performance on the mortality prediction task under different experimental conditions showed that supervision was more likely to improve performance. To further improve performance, we modified the algorithm to incorporate the use of other variables not captured in the tensor during the tensor factorization and final model selection. These additional covariates improved the average AUC of the final models, and the combined use of covariates and computational phenotypes often outperformed either alone. The utility of incorporating SDOH confounding variables can be achieved by extending "confounding-aware" non-negative matrix factorization [23] to tensor factorization [7,28]. Similarly, the use of static variables alongside temporal variables in Afshar et al's 2019 TASTE tensor factorization model was shown to improve predictive power [24].

However, even with additional covariates, the performance of Kim et al.'s algorithm on the mortality prediction task ranged from poor to fair/good, depending on the cancer cohort. Additional features, especially known prognostic factors, from past medical history or pathology reports, can be easily added with the modifications we have made. While the predictive ability did not appear to correlate with the data quantity (e.g. total co-occurrence count of the tensor, or number of co-occurrences per patients), it may be affected by the data quality. Many patients originally identified by their index cancer were not included in the final tensor, due to the lack of any other encounter information. Some patients had many non-cancer-related medications related to multiple comorbidities but only one diagnosis – their index cancer – and were thus excluded due to indication filtering. These scenario suggest that the EHR data was only able to capture a partial record of these patients' interactions with the health system. Data validation is also an important consideration; inaccuracies in diagnostic classification have been shown in the EHR data [25].

While the breast cancer cohort from the NMEDW has been used to evaluate machine learning methods on another classification task, identifying recurrences, those methods did not include using tensor factorization to derive features [26-28]. Future work with this data source may include comparing the discriminativeness of computational phenotypes versus simpler features. One can also explore using different classification algorithms both to guide tensor factorization and in the final classification step which uses the discovered computational phenotypes. In this paper, the application of tensor factorization to cancer cohorts in the NMEDW, and the evaluation of the resulting computational phenotypes and their ability to predict mortality, thus establish a useful baseline for future work with this data source.

**Conclusions**

Constrained tensor factorization was applied to EHR data of patients with cancer in the Northwestern Medicine Enterprise Data Warehouse to yield computational phenotypes that are concise and useful for mortality prediction. Restricting tensor entries to only those which are medically indicated greatly improved phenotype conciseness and ease of summarization, and the use of SDOH covariates during factorization and prediction consistently improved predictive accuracy.

**Declarations**

Ethics approval and consent to participate: The study was reviewed and approved by the Northwestern University Institutional Review Board (IRB ID STU00202167). Waiver of HIPAA authorization and consent process was obtained.

Consent for publication: Not applicable.

Availability of data and materials: The datasets generated and analyzed during the current study are not publicly available due to Northwestern Medicine Enterprise Data Warehouse policies.

Competing Interests: The authors declare that they have no competing interests.

Funding: The study is supported in part by NIH grant 1R01LM013337.

Author's contributions: FYC processed and analyzed the patient data, and drafted the manuscript. CM obtained the patient data and revised the manuscript. YL conceptualized the study, provided guidance on data analysis, and revised the manuscript. All authors read and approved the final manuscript.

Acknowledgements: None

Author's information: None

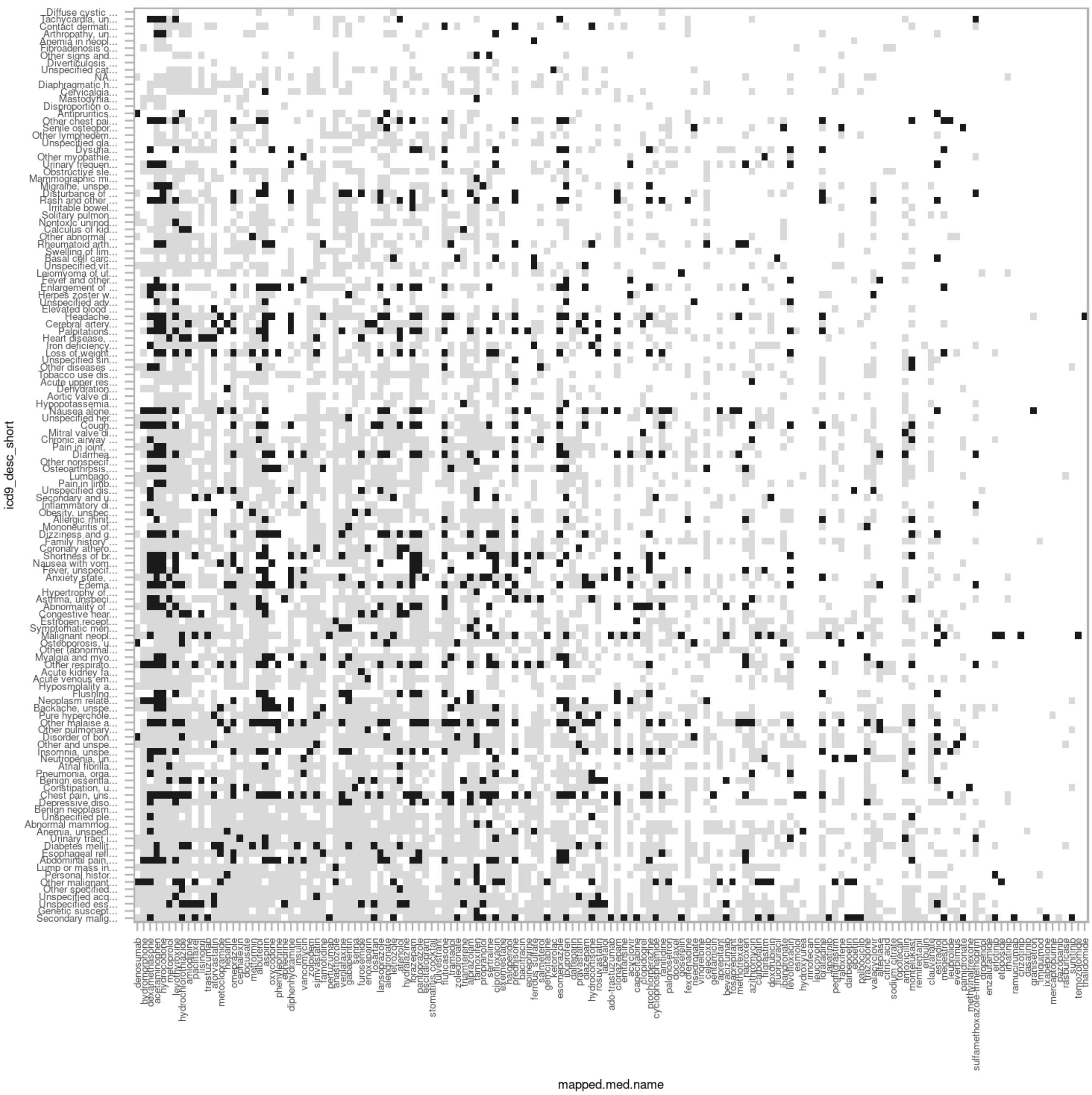

**Figure S1.** Diagnosis-medication pairs before and after indication filtering. Filled squares indicate that a diagnosis-medication pair was observed in the same patient encounter. Indicated pairs are shown in black; non-indicated are shown in grey. This plot is based on data from the breast cancer cohort.

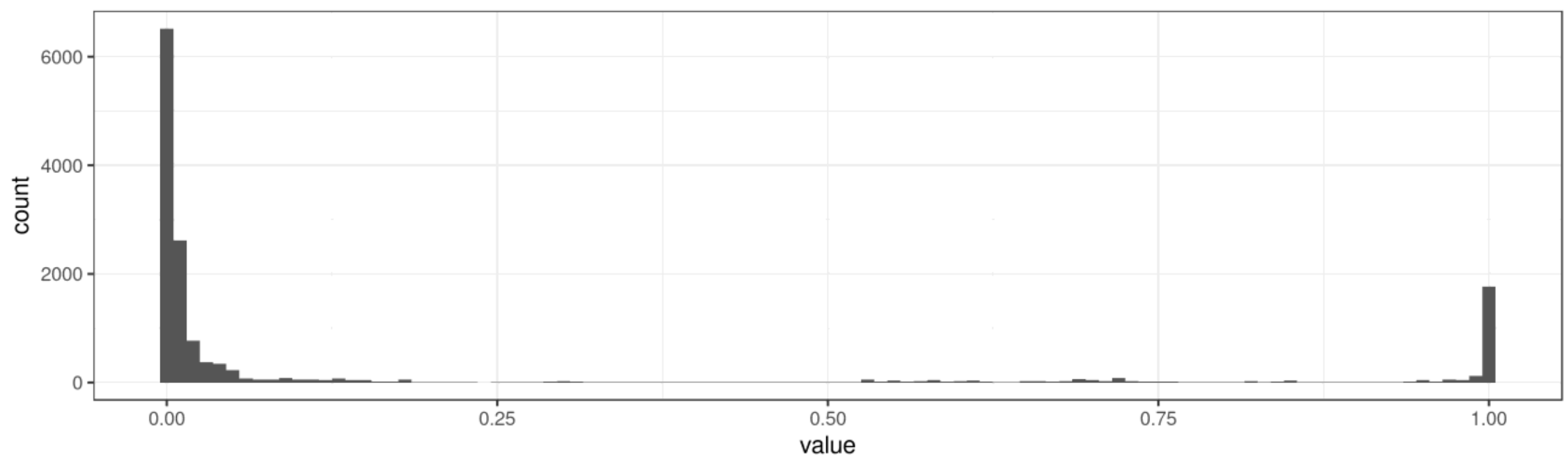

**Figure S2.** Membership value distribution for diagnoses. The distribution of membership values for diagnoses, across all phenotypes from all cross-validation folds for the breast cancer cohort. Tensor factorization was performed with a supervised term, indication filtering, and SDOH covariates.

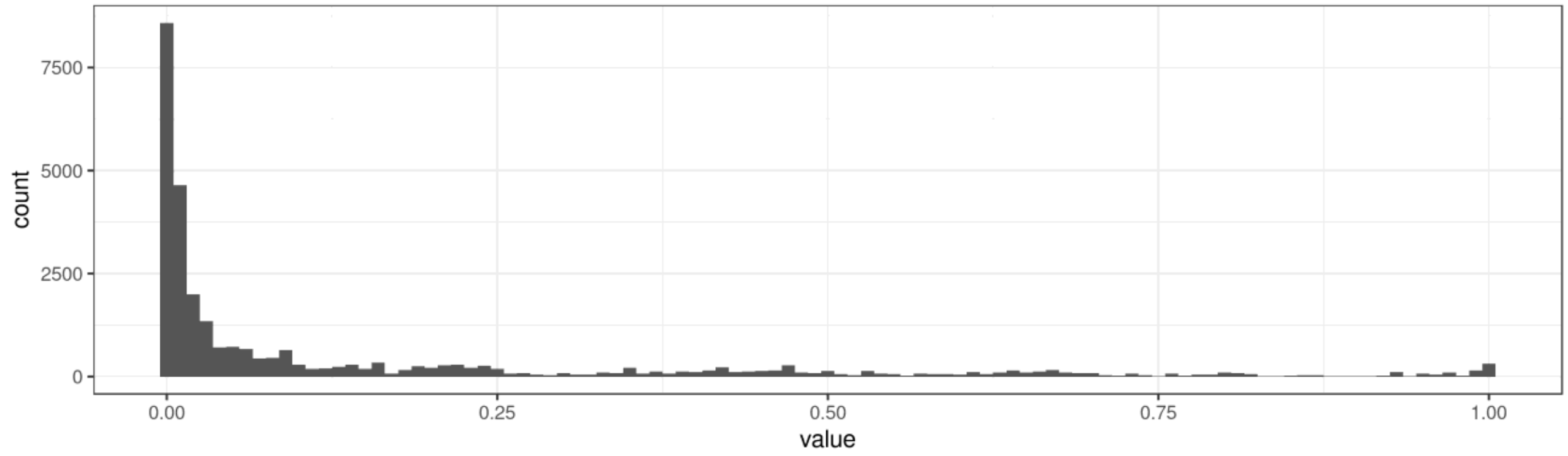

**Figure S3.** Membership value distributions for medications. The distribution of membership values for medications, across all phenotypes from all cross-validation folds for the breast cancer cohort. Tensor factorization was performed with a supervised term, indication filtering, and SDOH covariates.

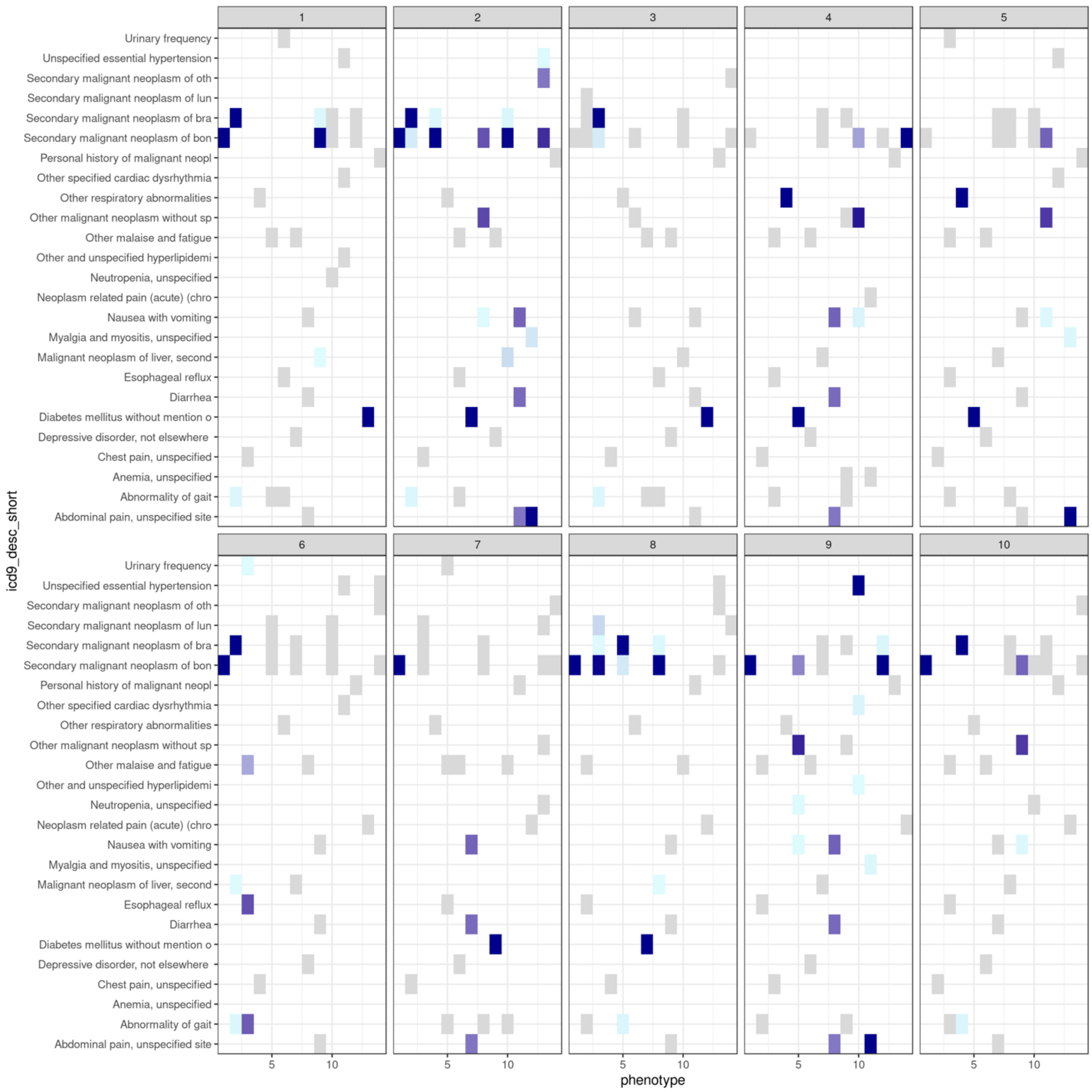

**Figure S4.** Diagnoses represented in the top 15 phenotypes for breast cancer cohort. Each panel shows the presence (filled rectangle) or absence of common diagnoses (y-axis) in the first 15 computational phenotypes (x-axis) for one cross-validation fold. Phenotypes not selected to the final model by stepwise regression are have grey rectangles in that column; those that are selected are shown in color, with darker colors representing a larger membership value. This plot is based on tensor factorization of the breast cancer tensor with a supervised term, indication filtering, and SDOH covariates.